\documentclass[runningheads]{llncs}
\usepackage[T1]{fontenc}
\usepackage{graphicx}
\usepackage{subfig}
\usepackage{multirow}
\usepackage{multicol}
\usepackage{amssymb}
\usepackage{amsmath}
\usepackage{makecell}
\usepackage{svg}
\usepackage{booktabs}
\usepackage{cite}
\usepackage{multicol}
\usepackage{marvosym}\usepackage{pgfplots}
\pgfplotsset{compat=1.18}

\begin{document}

\title{Label-template based Few-Shot Text Classification with Contrastive Learning}
\author{Guanghua Hou \and
Shuhui Cao \and
Deqiang Ouyang \and
Ning Wang\thanks{Corresponding author}
}
\authorrunning{G. Hou et al.}
\titlerunning{Label-template based Few-Shot Text Classification}
\institute{School of Computer Science, Chongqing University, Chongqing, China
\email{hgh@stu.cqu.edu.cn,\{csh17, deqiangouyang, nwang5\}@cqu.edu.cn}}
\maketitle \begin{abstract}
As an algorithmic framework for learning to learn, meta-learning provides a promising solution for few-shot text classification. 
However, most existing research fail to give enough attention to class labels. 
Traditional basic framework building meta-learner based on prototype networks heavily relies on inter-class variance, and it is easily influenced by noise.
To address these limitations, we proposes a simple and effective few-shot text classification framework.
In particular, the corresponding label templates are embed into input sentences to fully utilize the potential value of class labels, guiding the pre-trained model to generate more discriminative text representations through the semantic information conveyed by labels. 
With the continuous influence of label semantics, supervised contrastive learning is utilized to model the interaction information between support samples and query samples. 
Furthermore, the averaging mechanism is replaced with an attention mechanism to highlight vital semantic information.
To verify the proposed scheme, four typical datasets are employed to assess the performance of different methods.
Experimental results demonstrate that our method achieves substantial performance enhancements and outperforms existing state-of-the-art models on few-shot text classification tasks.
\end{abstract}
%第一页会议信息（可删）
% \renewcommand{\headrulewidth}{0pt}
% \renewcommand{\footrulewidth}{0pt}
% \thispagestyle{fancy}
% \fancyhead{}
% \chead{\small 2024 International Conference on Neural Information Processing (ICONIP)}
% \setlength{\headheight}{15pt}  % 增加页眉的高度（如果需要）
% \setlength{\headsep}{0.5 cm}  % 调整页眉与正文之间的间距

\section{Introduction}
Text classification is a cornerstone in Natural Language Processing (NLP)~\cite{liSurveyTextClassification2022,tsirmpasNeuralNaturalLanguage2024}. 
Traditional deep learning approaches for text classification heavily rely on large-scale labeled data to achieve satisfactory generalization performance on unseen datasets \cite{devlinBERTPretrainingDeep2019,johnsonDeepPyramidConvolutional2017}. 
However, in many real-world scenarios, such as medicine, finance, and biology, obtaining sufficient labeled data is challenging due to privacy concerns and costs. Unlike conventional machine learning systems, humans possess a remarkable ability to rapidly acquire new knowledge and establish cognition through limited examples. 
For instance, a child can master the names and traits of various animals and accurately differentiate them after viewing only a small number of animal images. The rapid learning capability has given rise to the field of Few-Shot Learning (FSL).
\\ \indent Few-Shot Text Classification (FSTC) is an important application of FSL in the text domain, which emphasizes building high-performance classifier models with minimal labeled data. 
In recent years, this research topic has garnered significant attention due to its practical relevance and application potential. 
In general, Meta-learning has emerged as a dominant paradigm for FSTC, enabling the transfer of knowledge learned from source classes with abundant labeled data to target classes unseen during training \cite{finnModelAgnosticMetaLearningFast2017,snellPrototypicalNetworksFewshot2017,sungLearningCompareRelation2018,liuBoostingFewshotText2023,hanMetaLearningAdversarialDomain2021}. 
A leading approach in meta-learning approaches is the prototypical network \cite{snellPrototypicalNetworksFewshot2017}. 
Increasing studies have adopted it as a foundational framework, enhancing meta-learning in FSTC with new innovations and optimizations.
\\ \indent
However, most existing studies based on prototypical networks have not given sufficient attention to class labels, where the label information is merely utilized in the final classification step, leading to a waste of label semantics. 
For example, by adding class labels to the input sentences, the feature extraction capability of pre-trained models can be effectively improved in the scenarios with scarce data \cite{luoDonMissLabels2021, wangJointFewShotText2022}. 
Unfortunately, labels are not structurally complete, independent semantic units, and directly connecting these labels to sentences often leads to a lack of fluency and semantic coherence in the text. 
This unnatural form of text will significantly influence the feature extraction process of pre-trained models.
Moreover, existing meta-learners based on prototype networks usually calculate separate prototypes for each class using only support samples, without fully exploiting the intra-class and inter-class relationships among the samples \cite{snellPrototypicalNetworksFewshot2017,luoDonMissLabels2021,sunMEDAMetaLearningData2021,pengFewshotTextClassification2023}. Since these methods exhibit a high degree of dependence on the inter-class variance of the sample data, they will struggle to make accurate predictions when the target data to be classified comprises relatively similar categories.
Furthermore, prototype networks typically simply average the representations within the same class to generate class prototypes. This averaging mechanism is highly susceptible to the negative impact of individual sample biases with the limited number of samples per class, rendering the class prototypes sensitive to outliers.
\\ \indent
To address the issues mentioned above, this paper presents a unique FSTC framework that ingeniously integrated label semantics, contrastive learning, and attention mechanisms. 
Specifically, the corresponding label templates is embed into the input sentences to fully leverage the potential value of class labels. 
Thus, the semantic information conveyed by the labels can guide the pre-trained model to capture more discriminative text embeddings.
Furthermore, to capture subtle differences between various categories, we employ supervised contrastive learning which pulls text representations of the same class closer in the feature space and push apart representations of different categories.
Besides, an attention mechanism is introduced into the prototype network, replacing the simple averaging mechanism with an attention mechanism to highlight important semantic information. 
This improved mechanism will enable the prototype network to generate more representative and discriminative class prototypes. In our approach, label semantics are employed not only to improve feature extraction but also to enhance contrastive learning, allowing it to identify more similar and distinct features across various samples.
Meanwhile, label semantics can also assist the attention-based prototype network in achieving superior predictive performance on test data.
Our method has been shown to be effective through experimentation. Moreover, compared to existing schemes, our model converges faster, requiring only a few iterations during training to achieve high classification accuracy.

The key contributions of this paper are outlined as follows:
\renewcommand{\labelitemi}{\textbullet}
\begin{itemize}
    \item We introduce and implement a simple yet effective FSTC framework which integrates label templates, contrastive learning, and attention mechanisms. 
    \item We perform a thorough analysis examining how different components of our method impact FSTC tasks, highlighting the significance of label semantics.
    \item Comprehensive experiments conducted on four public benchmark datasets demonstrate that our method remarkably surpasses existing strong baseline models in FSTC tasks, especially in one-shot scenario.
\end{itemize}

\section{Related Work}
\subsection{Fine-tuning Based approaches}
The Fine-tuning method utilizes a pre-trained model on source classes as a starting point and adapts to new tasks by making minor parameter adjustments on target classes. This method fully leverages the rich knowledge from source classes while avoiding the dilemma of training the model from scratch on target classes. The key to the fine-tuning method lies in how to effectively utilize the knowledge from the pre-trained model and find the most suitable parameter configuration for the characteristics of the target task. Howard et al. \cite{howardUniversalLanguageModel2018} proposed ULMFiT, a universal language fine-tuning model aimed at being applicable to all NLP tasks. Nakamura and Harada \cite{nakamuraRevisitingFinetuningFewshot2019} demonstrated that employing an adaptive gradient optimizer during fine-tuning enhances test accuracy. Suchin et al. \cite{gururanganDonStopPretraining2020} performed a second pre-training step on the pre-trained model before fine-tuning, including domain-adaptive and task-adaptive pre-training. 

Recently, innovative prompt-based methods have been presented \cite{wangEntailmentFewShotLearner2021,xieLeveragingInterclassDifferences2023}. These methods introduce additional prompt information into the model's input to guide the learning process and have shown promising performance in FSTC tasks. Prompts typically consist of selected words, phrases, or sentences that provide context and semantic information for the target task \cite{liuPretrainPromptPredict2023}. However, when the source and target datasets exhibit significant distribution shifts, fine-tuning based methods face the challenge of over-fitting in few-shot scenarios with scarce data. 

\subsection{Meta-learning Based approaches}
Meta-learning intends to build models that can rapidly acclimate to target tasks of unseen classes with some small tasks constructed by a few samples of seen classes during training. Current research on meta-learning generally falls into two categories: optimization-based approach and metric-based approach. Optimization-based meta-learning concentrates on learning to update model parameters via gradient descent from a limited number of labeled training examples, highlighting the model's intrinsic parameter update mechanism.
Representative methods include Meta-Learner MAML \cite{finnModelAgnosticMetaLearningFast2017}, Meta-SGD \cite{liMetaSGDLearningLearn2017}, and Reptile \cite{nicholFirstOrderMetaLearningAlgorithms2018}. Metric-based meta-learning, such as matching networks \cite{vinyalsMatchingNetworksOne2016}, prototypical networks \cite{snellPrototypicalNetworksFewshot2017}, relation networks \cite{sungLearningCompareRelation2018}  and inductive networks \cite{gengInductionNetworksFewShot2019}, aim to learn a metric space capable of assessing the similarity between samples.

Our method is also a metric-based meta-learning approach. Its core lies in the ingenious integration of label templates, contrastive learning, and attention mechanism.
Remarkably, in our approach, label templates not only enhance the feature extraction process, but also further strengthen the learning capabilities of contrastive learning and attention-based prototypical networks.
\section{Problem Formulation}
In this chapter, we will present a summary of the standard meta-learning architecture and describe the the problem formulation of FSTC tasks. Formally, let $\mathcal{Y}_{train}$, $\mathcal{Y}_{valid}$, and $\mathcal{Y}_{test}$ represent the Non-overlapping sets of training, validation, test classes and use $\mathcal{S}$, $\mathcal{Q}$ as support set and query set. 

\noindent
\textbf{Training Phase}
For a $\mathbf{n}$-way $\mathbf{k}$-shot text classification problem, at each episode, we first sample $\mathbf{n}$ classes from $\mathcal{Y}_{train}$. For each class, $\mathbf{k}$ instances are sampled to form $\mathcal{S}$, and another disjoint $\mathbf{m}$ instances are sampled to form $\mathcal{Q}$. We use $(x_i^s, y_i^s)$ to denote the $i$-th instance in $\mathcal{S}$, where $x_i^s$ is the input text and $y_i^s$ is the corresponding label. Respectively, $(x_j^q, y_j^q)$ is used to denote the $j$-th instance in $\mathcal{Q}$. The model's goal during training is to learn how to classify $\mathcal{Q}$ based on $\mathcal{S}$.

\noindent
\textbf{Testing Phase}
After training, the model is evaluated on $\mathcal{Y}_{valid}$ and $\mathcal{Y}_{test}$, where the classes are unseen to those in $\mathcal{Y}_{train}$. The evaluation process is similar to the training phase: for each episode, $\mathbf{n}$ classes are sampled from $\mathcal{Y}{valid}$ or $\mathcal{Y}{test}$, and $\mathbf{k}$ instances per class are sampled to form $\mathcal{S}$, with another disjoint $\mathbf{m}$ instances forming $\mathcal{Q}$. The model's performance is assessed based on its ability to classify $\mathcal{Q}$ using $\mathcal{S}$.

\section{Method}
\begin{figure}[t]
    \centering
    \includegraphics[scale=0.13]{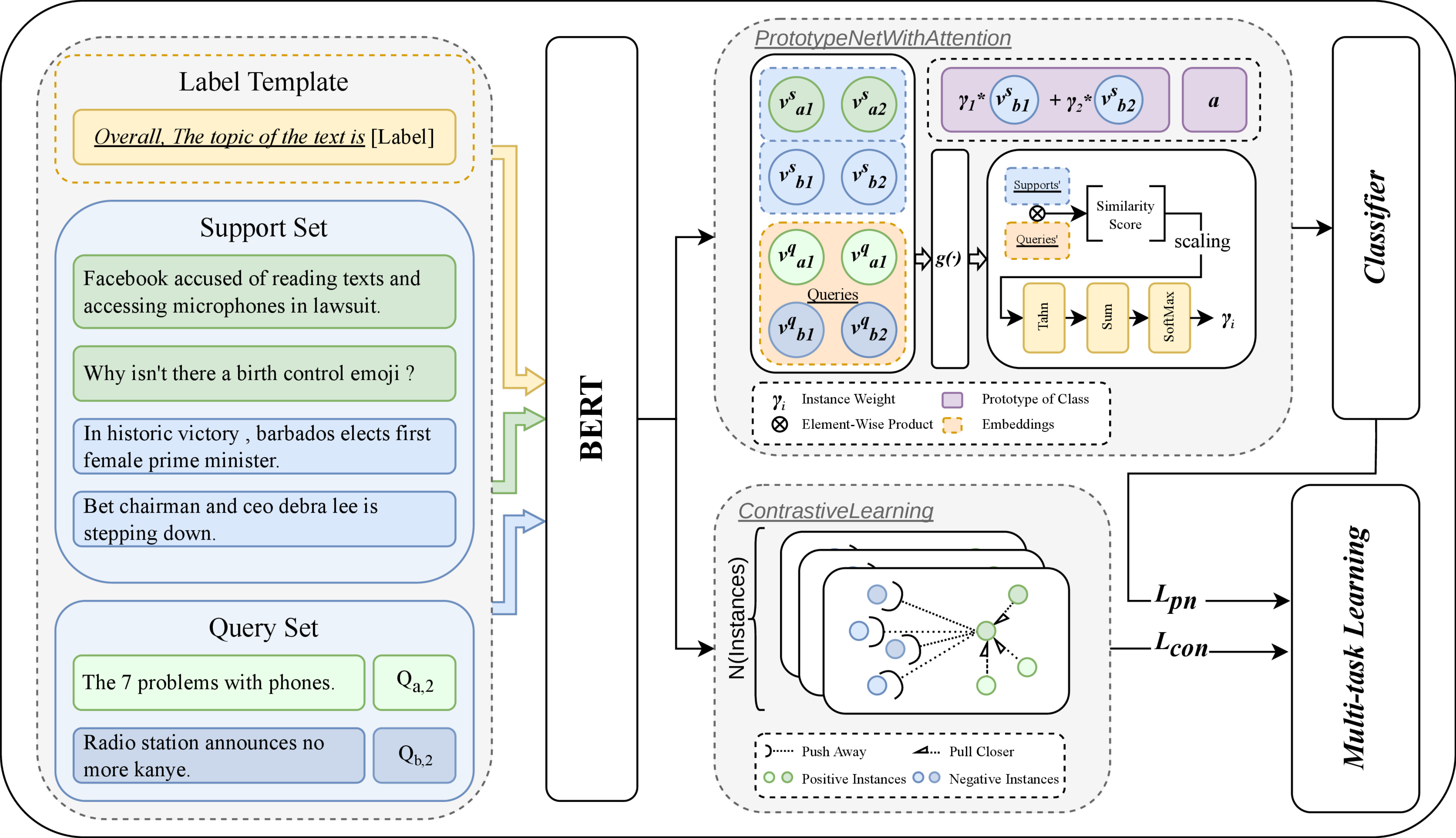}
    \caption{The structure of our method. Eight instances from two classes $a$ and $b$ are sampled to form  $\mathcal{S}$ and $\mathcal{Q}$. ${v}^{{s,q}}_{ci}$ is the text representation. In the prototype network, a fully connected layer $g(\cdot)$ produces the attention-awared representation ${{v}'}$. The weight $\gamma_i$ is applied to the $i$-th representation to obtain the class prototype for classification. $L_{pn}$ and $L_{con}$ are used for multi-task learning.}
    \label{plot:model_structure}
\end{figure}

\subsection{Feature Extraction with Label Information}
In this work, We transform labels by utilizing artificially constructed templates. First, each label is expressed as a structurally complete sentence, and then it is concatenated with the original text to form a coherent text that simultaneously contains sentence semantics and label semantics. We design a series of label templates by adding prompt words. After repeated experiments, "\textbf{\textit{Overall, the topic of the text is}}" performs the best in various datasets. Therefore, in subsequent experiments, we will adopt this template as the final label template $T$.

Recently, pre-trained language models have shown remarkable performance in feature extraction and representation learning. In our work, following the previous works in FSTC \cite{luoDonMissLabels2021,chenContrastNetContrastiveLearning2022}, we utilize pre-trained BERT as text encoder $f_\theta(\cdot)$ to project the input text into a high-dimensional feature space. Specifically, BERT processes the input text combined with the label template $(x_c,T_c)$, which is tokenized for analysis. We extract feature vectors from \textit{CLS} token as the text representation for the input text sequence. 
\noindent
\begin{equation}
    v=f_\theta(x_c ,l_c), v\in R^d,
\end{equation}
\begin{equation}
    l_c=T_c:{label}_c,
\end{equation}
where $\theta$ is the parameter and $v$ is the output of BERT, $l_c$ and ${label}_c$ are the concatenated label template and label text that from the $c$-th class. "${:}$" is the concatenating operation.

When there are limited training data, the label template can effectively enhance the feature extraction process of BERT, guiding it to generate more discriminative text embeddings that is related to the class semantic. It is noteworthy that support instances and query instances serve different purposes, necessitating distinct template processing approaches. For support samples, we concatenate the formatted label template with the original sentence to obtain a templated input text incorporating class name information. For query samples, since the true labels are unavailable, we retain the initial input format of each sentence without modification to ensure the normal progression of the model's prediction process. Therefore, for an $N$-way $K$-shot FSTC task, the support set $\mathcal{S}$ and query set $\mathcal{Q}$ within a episode can be represented as:

\begin{equation}
    \begin{aligned}
    & \mathcal{S} = \{(x_{c,i}^s,l_c),y_{c,i}^s\}_{c=1,i=1}^{N,K}, \mathcal{Q} = \{x_{c,j}^q, y_{c,j}^q\}_{c=1,j=1}^{N,M},
    \end{aligned}
\end{equation}
where $x_{c,i}^s$ and $y_{c,i}^s$ represent the $i$-th support instance and its ground-truth label in the $c$-th class. $x_{c,j}^q$ and $y_{c,j}^q$ denote the j-th query instance and its label in the $c$-th class, respectively.

\subsection{Label-semantic Augmented Supervised Contrastive Learning}
To further improve the pre-trained model's capacity to utilize and integrate label semantics, we incorporate the idea of using label semantics to enhance feature extraction into the contrastive learning framework. Supervised contrastive learning is employed to bring text representations of the same class closer and push away those of different categories. Guided by enriched semantic information for class labels, contrastive learning can identify more similar features within the same class and more distinct features across different categories. 

Under $N$-way $K$-shot setting, given $\mathcal{S}$ and $\mathcal{Q}$ in an episode, we combine the $N * K$ support instances $\{(x_{c,i}^s, l_c),y_{c,i}^s\}$ and the $N * M$ query instances $\{x_{c,j}^q, y_{c,j}^q\}$ into a single set $\mathcal{D}=\{x_p,y_p\}_{p=1}^{N(K+M)}$, where the index of the instances is $I=\{1,2,...NK,...,N(K+M)\}$. It is a critical step to construct positive pairs and negative pairs for contrastive learning.
Following Khosla et al.'s supervised contrastive learning method \cite{khoslaSupervisedContrastiveLearning}, we construct positive pairs $(x,x^+)$ and negative pairs $(x,x^-)$ based on the instance label in one episode. 
Specifically, for each instance $x_c$ in the set $\mathcal{D}$, $(K+M-1)$ instances from the same class $c$ are selected as positive instances, and $(N-1)(K+M)$ instances from different classes are selected as negative instances. 
Considering instance $x_p$, the contrastive loss $L_{con}$ can be formulated as follows: 
\begin{equation}
    sim(v_1,v_2) = \frac{v_1\cdot v_2}{||v_1||\cdot ||v_2||},    
\end{equation}
\begin{equation}
L^p_{con} = -\frac{1}{|H(p)|} \log\frac{\sum_{h\in H(p)}exp(sim(v_p,v_h))/\tau}{\sum_{t\in A(p)}exp(sim(v_p,v_h))/\tau}, 
\end{equation}
\begin{equation}
H(p) = \{h\in A(p)|y_h=y_p\}, 
\end{equation}
\begin{equation}
A(p) = I \setminus {p},
\end{equation}
where $sim(v_1,v_2)$ is cosine similarity between two text representations $v_1$ and $v_2$, $L^p_{con}$ represents the contrastive learning loss for sample $p$. 
Let $A(p)$ denote the set of all samples excluding sample $p$, and let $H(p)$ represent the set of samples that share the same label as sample $p$.$\tau \in R^+$ represents the temperature factor used to adjust the scale of the similarity score.

For a batch $D=\{x_p,y_p\}_{p=1}^{N(K+M)}$, the overall contrastive loss $L_{con}$ is computed as the sum of the contrastive loss for each instance in the batch: 
\begin{equation}
    L_{con} = \sum_{p=1}^{N(K+M)} L^p_{con}.
\end{equation}

The above formula calculates the similarity between pair of instances instead of class level, which effectively capture and model the relationships and inter-dependencies between any pair of instances. 
By incorporating label information into the supervised contrastive learning, the features extracted between positive and negative pairs become more representative. As a result, text representations within the same class become more concentrated in the feature space, while representations of various categories are more dispersed. 

\subsection{Update Prototype network through Attention Mechanism and Label-semantic}
The attention mechanism employed in this work is inspired by the HATT \cite{gaoHybridAttentionBasedPrototypical2019} model, we only use the instance-level attention and incorporate it into the prototype network.By considering the relevance between support samples and query samples, the attention mechanism assigns different weight parameters to each support sample. This allows the prototype network to focus more on support samples that are closely related to the query samples in specific aspects. As a result, when calculating class prototypes, these relevant support samples carry greater weight compared to others. Specifically, for a meta-learning task consisting of $n$ classes sampled using $N$-way $K$-shot strategy, we first calculate attention weights $\gamma_{i}^c$ of each support sample $(x_{c,i}^s,l_c)$ from class $c$ with respect to all query samples ${x^q}$ in the query set using the following formula:
\begin{equation}
\gamma_{i}^c = \frac{exp(e^c_i)}{\sum_{j=1}^{K}exp(e^c_j)},
\end{equation}
\begin{equation}
e^c_i = sum\{\sigma(g(v^s_{c,i})\odot g(v^q))\},
\end{equation}
\begin{equation}
v^s_{c,i} = f_\theta(x^s_{c,i},l_c) , v^q=f_\theta(x^q),
\end{equation}
where $v^s_{c,i}$ are the text representations of the $i$-th support sample in the $c$-th class. Respectively, $v^q \in \mathcal{Q}$ is from the same class, but the model is unaware of its class. $g(\cdot)$ constitutes a fully connected layer that maintains identical dimensions for both its input and output. $\odot$ is element-wise multiplication, and $\sigma$ is $tanh$ activation function.

Then, the class prototype $w_c$ of any given class $c$ will be expressed as the weighted average sum of the support samples. Given a query sample $x^q$ to be classified, we can get its probability of belonging to each class by calculating the cosine similarity with the class prototypes $w_c$.
The process can be described as:
\begin{equation}
w_c=\sum_{(x_{c,i}^s,l_c)\in \mathcal{S}_c}\gamma_{i}^c\cdot f_{\theta}(x^s_{c,i},l_c), 
\end{equation}
\begin{equation}
P(c|x^q)=\frac{exp(-d(f(x^q),w_c))}{\sum_{\widetilde{c}=1}^{N}exp(-d(f(x^q),w_{\widetilde{c}})),}
\end{equation}
where $P(c|x^q)$ represents the probability that sample $p$ belongs to category $c$, $d(x,y)$ is the euclidean distance function, which calculates the distance between $x$ and $y$ in the feature space. The objective function of the prototype network \cite{snellPrototypicalNetworksFewshot2017} is implemented through cross-entropy loss, with the specific formula:
\begin{equation}
    L_{pn}=-\sum_{q=1}^{NM}\sum_{c=1}^{N}y_{qc}\log P(c|x^q),
\end{equation}
where only when $x^q$ belongs to the $c$-th class, $y_{qc}$ is 1, otherwise 0.

\subsection{Objective Function}
Our work uses multi-task learning to optimize the model, addressing the overfitting issue common in FSL. To mitigate this, we introduce two tasks during training: contrastive learning and prototype learning. The contrastive learning task leverages supervised contrastive learning to obtain more discriminative feature representations. 
The goal is to cluster text representations of the same class while separating those of different classes. Meanwhile, the prototype classification task uses an attention-based prototype network to create more representative prototype representations for each class. The model is guided to optimize its learning towards the correct class for the query text.
The objective function $L$ is defined as:
\begin{equation}
    L=(1-\rho)L_{pn}+\rho L_{con},
\end{equation}
where $\rho$ serves as the weight parameter that balances the two loss. In our work, it is fundamental to find a proper $\rho$ and the changing strategy during the training process to balance the two tasks. Our optimized settings are presented in the experiment section.

\section{Experiment}
\subsection{Datasets}
Following ContrastNet \cite{chenContrastNetContrastiveLearning2022}, our experiments are conducted on 4 news and review datasets that are widely used in FSTC tasks. Details are shown in Table \ref{table:datasets}.
\renewcommand\arraystretch{0.9}
\begin{table}[t]
    \centering
    \caption{The statistics of four datasets. "Instance per Class" denotes the number of instances per class, "Samples" refers to the total count of samples, "Train/Valid/Test" denotes the number of classes present in the training, validation, and test sets, and "Avg.Length" denotes the average sentence length.}
    \setlength{\tabcolsep}{1.5mm}
    \begin{tabular}{ccccc}
    \toprule
    
    \textbf{Dataset} & \textbf{Instance per Class} & \textbf{Samples} & \textbf{Train/Valid/Test} & \textbf{Avg.Length} \\ 
    
    \midrule
    20News & 941 & 18820 & 8/5/7 & 340 \\ 
    Amazon & 1000 & 24000 & 10/5/9 & 140 \\ 
    HuffPost & 900 & 16900 & 20/5/16 & 11 \\ 
    Reuters & 20 & 620 & 15/5/11 & 168 \\ 
    \bottomrule
    \end{tabular}
    \label{table:datasets}
    \end{table}

\noindent
\textbf{20News} \cite{langNewsWeederLearningFilter1995} comprises 18,820 documents that originate from 20 different newsgroups. The average length of sentences is the longest among our 4 datasets. 

\noindent
\textbf{Amazon} \cite{heUpsDownsModeling2016} is composed of 142.8 million reviews across 24 categories. We adhere to the setting in \cite{hanMetaLearningAdversarialDomain2021} which only uses a subset containing 1000 reviews per class.

\noindent
\textbf{HuffPost} \cite{baoFewshotTextClassification2020} is a dataset of news articles from the HuffPost website. It contains 900 samples per class and 41 classes.

\noindent
\textbf{Reuters} \cite{baoFewshotTextClassification2020} is a news dataset sourced from Reuters newswire in 1987. We follow \cite{baoFewshotTextClassification2020} and only use 31 classes with 20 samples per class.

\subsection{Baselines}
We evaluate our method against the following baselines:

\noindent
\textbf{PN} \cite{snellPrototypicalNetworksFewshot2017}(Prototype Network) is a straightforward and effective metric-based FSL method that aligns query instances with class prototypes.

\noindent
\textbf{MAML} \cite{finnModelAgnosticMetaLearningFast2017} is an optimization-based approach that enables rapid acclimatization to new tasks through just several gradient updates.   

\noindent
\textbf{Ind} \cite{gengInductionNetworksFewShot2019}(Induction Network) introduces a dynamic routing algorithm to learn class-level representaions, which shows strong performance in FSTC tasks.

\noindent
\textbf{DS-FSL} \cite{baoFewshotTextClassification2020} uses a MLP(Multilayer perceptron) in Prototypical Network and extends the model. It also builds a meta-learner features generalization ability. 

\noindent
\textbf{MLADA} \cite{hanMetaLearningAdversarialDomain2021} introduces an adversarial network to heighten the cross-domain transferability of meta-learning and create superior sentence embeddings. 

\noindent
\textbf{LaSAML} \cite{luoDonMissLabels2021} is a method that incorporates the label information in FSL to produce discriminative feature of input texts by adding label token after the text token.

\noindent
\textbf{ContrastNet} \cite{chenContrastNetContrastiveLearning2022} is a supervised contrastive learning model that involves both the support and query instances.  

\subsection{Implement Details}
The model undergoes training and evaluation on standard 5-way 1-shot and 5-way 5-shot FSTC tasks across multiple datasets. 
Consistent hyperparameters and training methodologies are applied to all datasets, following the approach outlined by \cite{chenContrastNetContrastiveLearning2022}. 

We use Adam optimizer at the learning rate of 1e-6 and an early stopping strategy is implemented with a patience of 3 epochs, activated if validation accuracy fails to improve. Training is capped at 10,000 iterations, with evaluations every 100 iterations. Validation and test results are averaged over 1,000 episodes. 5-fold cross-validation is employed to enhance evaluation. The max sequence length is 256. The temperature factor $\tau$ for contrastive learning is set to 5, and the loss weight $\rho$ increases linearly from 0 to 1 during training.

\subsection{Experiment Results}
\renewcommand\arraystretch{0.8}
\begin{table}[t]
    \centering
    \caption{The accuracy and macro-f1 score results on four text datasets. Results marked with "*" are sourced from \cite{liuLiberatingSeenClasses2024}, which has similar settings to our experiments.}
    \setlength{\tabcolsep}{1.3mm}
    \begin{tabular}{@{}l|cccc|cccc@{}}
        \toprule
        \multirow{4}{*}{\textbf{Method}} & \multicolumn{4}{c|}{20News} & \multicolumn{4}{c}{Amazon} \\ \cmidrule(l){2-9} 
                          & \multicolumn{2}{c|}{1-shot} & \multicolumn{2}{c|}{5-shot} & \multicolumn{2}{c|}{1-shot} & \multicolumn{2}{c}{5-shot} \\ \cmidrule(l){2-9} 
                          & Acc & F1 & Acc & F1 & Acc & F1 & Acc & F1 \\ \midrule
        PN*               & 0.4721 & 0.4426 & 0.6014 & 0.5948 & 0.5407 & 0.5320 & 0.6982 & 0.6979 \\
        MAML*             & 0.4309 & 0.4042 & 0.6194 & 0.6095 & 0.4900 & 0.4746 & 0.6987 & 0.6945 \\
        Ind*              & 0.4138 & 0.3715 & 0.4922 & 0.4910 & 0.5222 & 0.5049 & 0.5646 & 0.5508 \\
        DS*               & 0.6019 & 0.5729 & 0.8064 & 0.8011 & 0.6592 & 0.6434 & 0.8454 & 0.8422 \\
        MLADA*            & 0.6040 & 0.5776 & 0.7794 & 0.7752 & 0.6328 & 0.6143 & 0.8234 & 0.8184 \\
        ContrastNet       & 0.7204 & 0.6568 & 0.8062 & 0.7977 & 0.7536 & 0.6951 & 0.8439 & 0.8377 \\
        LaSAML            & 0.6872 & 0.6175 & 0.7871 & 0.7702 & 0.7566 & 0.6981 & 0.8397 & 0.8282 \\
        \midrule
        \textbf{Ours}     & \textbf{0.7486} & \textbf{0.6884} & \textbf{0.8333} & \textbf{0.8226} & \textbf{0.8021} & \textbf{0.7508} & \textbf{0.8618} & \textbf{0.8532} \\ 
        \midrule
        
        \multirow{4}{*}{\textbf{Method}} & \multicolumn{4}{c|}{HuffPost} & \multicolumn{4}{c}{Reuters} \\ \cmidrule(l){2-9} 
                        & \multicolumn{2}{c|}{1-shot} & \multicolumn{2}{c|}{5-shot} & \multicolumn{2}{c|}{1-shot} & \multicolumn{2}{c}{5-shot} \\ \cmidrule(l){2-9} 
                        & Acc & F1 & Acc & F1 & Acc & F1 & Acc & F1 \\ \midrule
        PN*               & 0.3436 & 0.3159 & 0.4940 & 0.4889 & 0.6233 & 0.6075 & 0.7188 & 0.7117 \\
        MAML*             & 0.2453 & 0.2349 & 0.3374 & 0.3284 & 0.5394 & 0.5275 & 0.7942 & 0.7900 \\
        Ind*              & 0.3525 & 0.3336 & 0.4341 & 0.4232 & 0.5578 & 0.5277 & 0.6425 & 0.6210 \\
        DS*               & 0.3946 & 0.3759 & 0.5986 & 0.5892 & 0.7572 & 0.7447 & \textbf{0.9404} & \textbf{0.9392} \\
        MLADA*            & 0.4433 & 0.4188 & 0.5736 & 0.5650 & 0.7178 & 0.7028 & 0.8252 & 0.8215 \\
        ContrastNet       & 0.5309 & 0.4532 & 0.6327 & 0.6204 & 0.8831 & 0.8495 & 0.9403 & 0.9378 \\
        LaSAML            & 0.5971 & 0.5184 & 0.6784 & 0.6618 & 0.8832 & 0.8501 & 0.9347 & 0.9309 \\ \midrule
        \textbf{Ours}     & \textbf{0.6725} & \textbf{0.6028} & \textbf{0.7176} & \textbf{0.7054} & \textbf{0.8892} & \textbf{0.8583} & 0.9302 & 0.9261 \\ 
        \bottomrule
    \end{tabular}
    \label{table:our_result}
    \end{table}

Table \ref{table:our_result} compares the performance of our model against the baselines across four datasets. Our model surpasses all baselines in both settings for four datasets, except in the Reuters 5-shot setting. The HuffPost appears to be the most challenging due to the short sentence length.

\noindent
\textbf{1-Shot Performance}
Our model surpasses all baselines in the 1-shot setting. It achieves an average accuracy improvement of 4.7\% and an average F1 improvement of 5.4\% comparing to the best baseline. On the most challenging dataset, our model shows the largest improvement, with a 7.5\% improvement in accuracy and an 8.4\% improvement in F1 measure.
The result reveals that our model effectively grasps the semantic context of the label and the text, especially when the length of the text is short.

\noindent
\textbf{5-Shot Performance}
Comparing to the best baseline, the model demonstrates an average accuracy improvement of 2.6\% and an average F1 improvement of 2.9\% in the 5-shot setting. In particular, our model also shows up to a 3.9\% improvement in accuracy and a 4.4\% improvement in F1 on the HuffPost. In the Reuters 5-shot setting, our model did not achieve optimal performance. This may be attributed to the characteristics of the Reuters dataset, which features a large number of categories with relatively few samples per category, resulting in our model not being fully adapted to this challenge.
\\ \indent The result indicate that our model achieve a more substantial boost in 1-shot than in 5-shot. It indicates that our model can generate more distinct class representations, particularly when the labeled class instances are scarce.

\subsection{Ablation Study}
We will investigate how components contribute to the overall method and demonstrate the effectiveness of each part individually and in combination in this section.
As shown in Table \ref{table:ourablation}, we achieve 8.12\% and 3.48\% improvement in 1-shot and 5-shot settings by means, compared to the Prototypical Network with BERT.
\begin{table}[h]
    \centering
    \caption{Performance of different classifiers in our model. The result is tested under the HuffPost's 5-shot setting.}
    \setlength{\tabcolsep}{3mm}
    \begin{tabular}{@{}ccc@{}}
      \toprule
      Method          & Acc             & F1              \\ \midrule
      KNN             & 0.6982          & 0.6892          \\
      PN              & 0.7168          & 0.7043          \\
      \textbf{PN-ATT} & \textbf{0.7176} & \textbf{0.7054} \\ \bottomrule
      \end{tabular}
    \label{table:classifiers}
  \end{table}
  
\noindent
\textbf{Effect of Attention Mechanism in Prototypical Network}
We denote the Prototypical Network with attention mechanism as $PN$-$ATT$. In the original method, prototypes are the mean of support instances, making the prototype a single instance in the 1-shot scenario. Consequently, in the ablation study shown in Table \ref{table:ourablation}, this improves the model's 5-shot performance but not the 1-shot setting. The attention mechanism helps produce more representative class prototypes, as shown in the Table \ref{table:classifiers} and \ref{table:ourablation}.

\noindent
\textbf{Effect of Contrastive Learning}
Contrastive learning helps the model to extract more robust and distinctive features from input texts by using all the instances in the episodes, which can make the prototype more representative. 
The result in Table \ref{table:ourablation} shows that it can improve the model's performance by 1.08\% and 1.66\% in 1-shot and 5-shot settings, respectively.

\renewcommand\arraystretch{0.8}
\begin{table}[t]
    \centering
    \caption{Ablation results of our methods, where "$\checkmark$" denotes the method is used in the model and "-"  denotes the method is not used. "AT", "CL", "LT" represent attention mechanism, contrastive learning, label template accordingly. The rightmost columns show the average relative changes of the model's performance in 1-shot and 5-shot settings.}
    \setlength{\tabcolsep}{0.29mm}
    \begin{tabular}{@{}ccccccccccccc@{}}
        \toprule
        \multicolumn{13}{c}{Accuracy}                                                                                                                                                                                                                                                                    \\ \midrule
        \multirow{2}{*}{AT} & \multirow{2}{*}{CL} & \multicolumn{1}{c|}{\multirow{2}{*}{LT}} & \multicolumn{2}{c|}{20News}       & \multicolumn{2}{c|}{Amazon}       & \multicolumn{2}{c|}{HuffPost}     & \multicolumn{2}{c|}{Reuters}                           & \multicolumn{2}{c}{Improvement}     \\ \cmidrule(l){4-13} 
                            &                     & \multicolumn{1}{c|}{}                    & 1-shot          & 5-shot          & 1-shot          & 5-shot          & 1-shot          & 5-shot          & 1-shot          & \multicolumn{1}{c|}{5-shot}          & \multicolumn{1}{c|}{1-shot} & 5-shot \\ \midrule
        -                  &-                   & \multicolumn{1}{c|}{-}                   & 0.6996          & 0.8086          & 0.7394          & 0.8007          & 0.4958          & 0.6642          & 0.8528          & \multicolumn{1}{c|}{0.9302}          & \multicolumn{1}{c|}{-}      & -      \\ \midrule
        -                   & -                   & \multicolumn{1}{c|}{$\checkmark$}        & 0.7345          & 0.8196          & 0.8006          & 0.8596          & 0.6597          & \textbf{0.7218} & 0.8816          & \multicolumn{1}{c|}{\textbf{0.9376}} & \multicolumn{1}{c|}{0.0722} & 0.0337 \\
        -                   & $\checkmark$        & \multicolumn{1}{c|}{-}                   & 0.7054          & 0.8188          & 0.7446          & 0.8483          & 0.5125          & 0.6738          & 0.8682          & \multicolumn{1}{c|}{0.9292}          & \multicolumn{1}{c|}{0.0108} & 0.0166 \\
        $\checkmark$        & -                   & \multicolumn{1}{c|}{-}                   & 0.6996          & 0.8120          & 0.7394          & 0.8511          & 0.4958          & 0.6675          & 0.8528          & \multicolumn{1}{c|}{0.9322}          & \multicolumn{1}{c|}{0}      & 0.0148 \\
        $\checkmark$        & $\checkmark$        & \multicolumn{1}{c|}{-}                   & 0.7074          & 0.8282          & 0.7448          & 0.8509          & 0.5118          & 0.6765          & 0.8593          & \multicolumn{1}{c|}{0.9263}          & \multicolumn{1}{c|}{0.0089} & 0.0245 \\ \midrule
        $\checkmark$        & $\checkmark$        & \multicolumn{1}{c|}{$\checkmark$}        & \textbf{0.7486} & \textbf{0.8333} & \textbf{0.8021} & \textbf{0.8618} & \textbf{0.6725} & 0.7176          & \textbf{0.8892} & \multicolumn{1}{c|}{0.9302}          & \multicolumn{1}{c|}{0.0812} & 0.0348 \\ \midrule
        \multicolumn{13}{c}{F1-measure}                                                                                                                                                                                                                                                                  \\ \midrule
        \multirow{2}{*}{AT} & \multirow{2}{*}{CL} & \multicolumn{1}{c|}{\multirow{2}{*}{LT}} & \multicolumn{2}{c|}{20News}       & \multicolumn{2}{c|}{Amazon}       & \multicolumn{2}{c|}{HuffPost}     & \multicolumn{2}{c|}{Reuters}                           & \multicolumn{2}{c}{Improvement}     \\ \cmidrule(l){4-13} 
                            &                     & \multicolumn{1}{c|}{}                    & 1-shot          & 5-shot          & 1-shot          & 5-shot          & 1-shot          & 5-shot          & 1-shot          & \multicolumn{1}{c|}{5-shot}          & \multicolumn{1}{c|}{1-shot} & 5-shot \\ \midrule
        -                   & -                   & \multicolumn{1}{c|}{-}                   & 0.6297          & 0.7927          & 0.6778          & 0.7942          & 0.4111          & 0.6642          & 0.8114          & \multicolumn{1}{c|}{0.9302}          & \multicolumn{1}{c|}{-}      & -      \\ \midrule
        -                   & -                   & \multicolumn{1}{c|}{$\checkmark$}        & 0.6714          & 0.8057          & 0.7494          & 0.8505          & 0.5862          & \textbf{0.7109} & 0.8480          & \multicolumn{1}{c|}{\textbf{0.9342}} & \multicolumn{1}{c|}{0.0812} & 0.0300 \\
        -                   & $\checkmark$        & \multicolumn{1}{c|}{-}                   & 0.6368          & 0.8049          & 0.6842          & 0.8382          & 0.4327          & 0.6566          & 0.8313          & \multicolumn{1}{c|}{0.9249}          & \multicolumn{1}{c|}{0.0137} & 0.0358 \\
        $\checkmark$        & -                   & \multicolumn{1}{c|}{-}                   & 0.6297          & 0.7967          & 0.6778          & 0.8410          & 0.4111          & 0.6488          & 0.8114          & \multicolumn{1}{c|}{0.9409}          & \multicolumn{1}{c|}{0}      & 0.0115 \\
        $\checkmark$        & $\checkmark$        & \multicolumn{1}{c|}{-}                   & 0.6384          & 0.8167          & 0.6843          & 0.8407          & 0.4314          & 0.6601          & 0.8201          & \multicolumn{1}{c|}{0.9221}          & \multicolumn{1}{c|}{0.0110} & 0.0145 \\ \midrule
        $\checkmark$        & $\checkmark$        & \multicolumn{1}{c|}{$\checkmark$}        & \textbf{0.6885} & \textbf{0.8226} & \textbf{0.7508} & \textbf{0.8532} & \textbf{0.6028} & 0.7054          & \textbf{0.8583} & \multicolumn{1}{c|}{0.9261}          & \multicolumn{1}{c|}{0.0926} & 0.0315 \\ \bottomrule
        \end{tabular}
    \label{table:ourablation}
    \end{table}

\noindent
\textbf{Effect of Label Template} 
Label Template shows the greatest improvement in both settings among three methods. By simply adding the label token after the text token, the model performance is dramatically improved, especially in the 1-shot scenario.
The efficacy of the Label Template technique underscores its promise as a strong strategy for increasing the accuracy and reliability of few-shot learning.
\\ \indent In addition, we note that in the 5-shot setting, the model that combines attention mechanisms, contrastive learning, and label templates does not surpass the model solely utilizing label templates in handling tasks from HuffPost and Reuters. This could be attributed to the former's more intricate structure and higher number of parameters, potentially increasing the likelihood of overfitting.

\subsection{Convergence Speed and Visualization}
\begin{figure}[h]
    \centering
    \subfloat[Loss]{\includegraphics[width=0.45\textwidth]{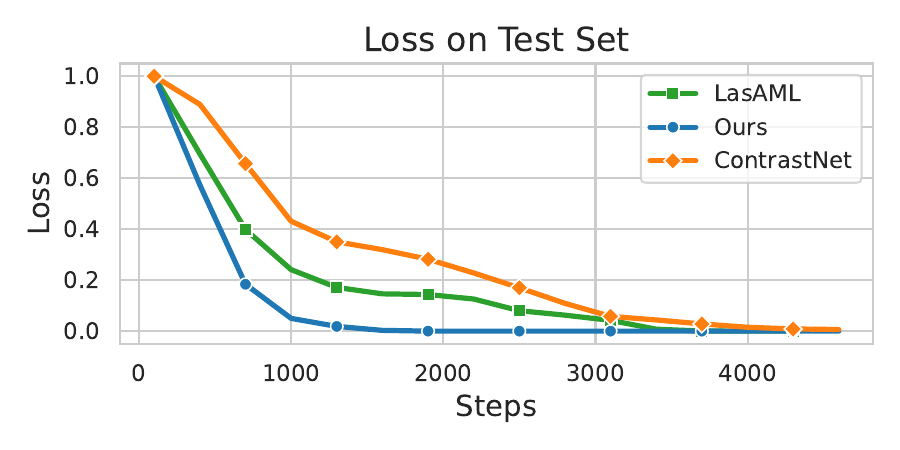}}
    \subfloat[Accuracy]{\includegraphics[width=0.45\textwidth]{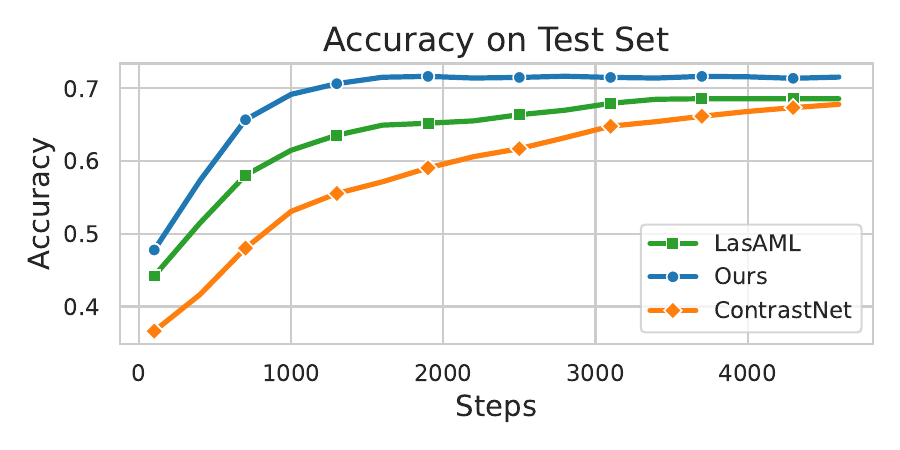}}
    \caption{The normalized loss and accuracy of different approach with learning rate of 1e-6 and early stop strategy. The results are averaged and sampled during the testing process of HuffPost dataset.}
    \label{figure:comparision_convergence_cn}
\end{figure}

\noindent
\textbf{Convergence Speed} We also compare the convergence rate of our model with two strong baselines on the HuffPost dataset. The figure \ref{figure:comparision_convergence_cn} shows that our method converges 2.62 times faster than ContrastNet under 5-shot setting with 8.49\% accuracy improvement. 

\pgfplotsset{width=6cm}

\begin{figure}[h]
    \centering
    \subfloat[ContrastNet]{\includegraphics[width=0.3\textwidth]{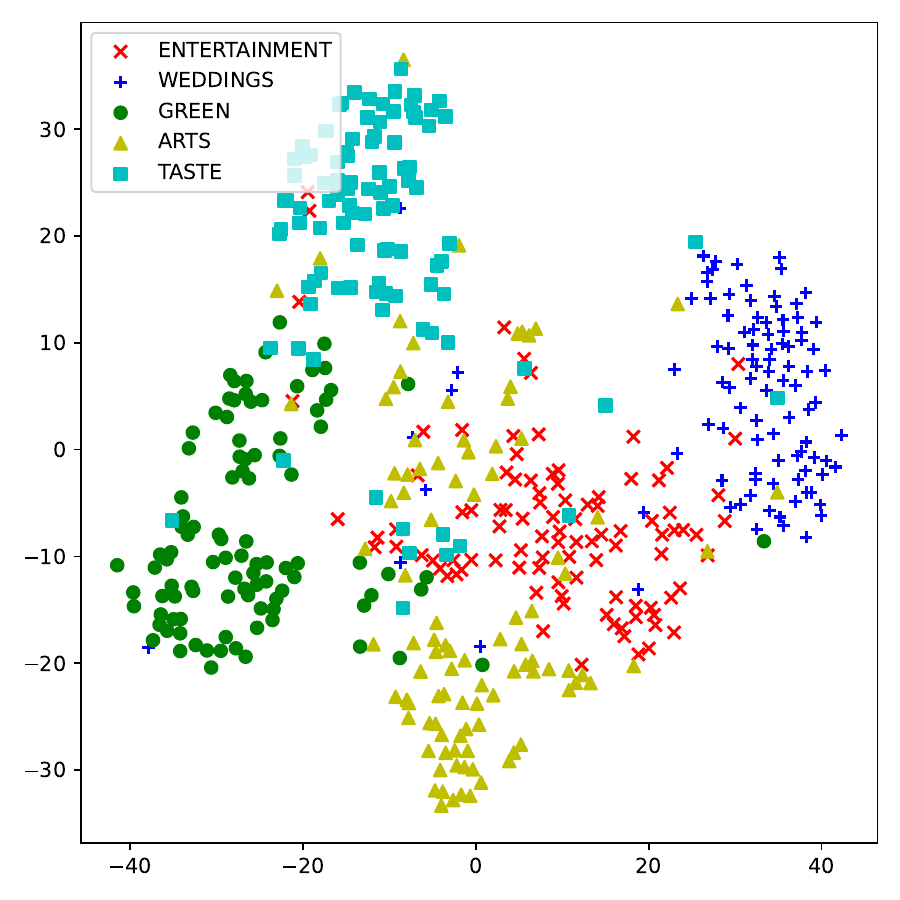}}
    \subfloat[LaSAML]{\includegraphics[width=0.3\textwidth]{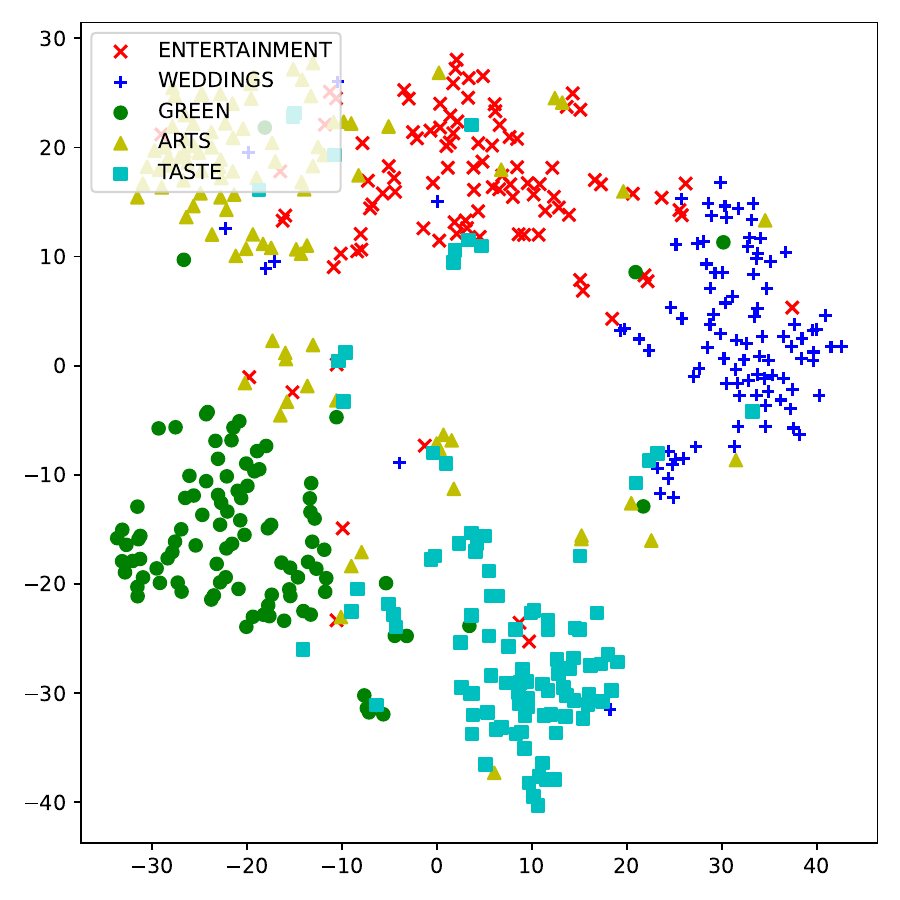}}
    \subfloat[Ours]{\includegraphics[width=0.3\textwidth]{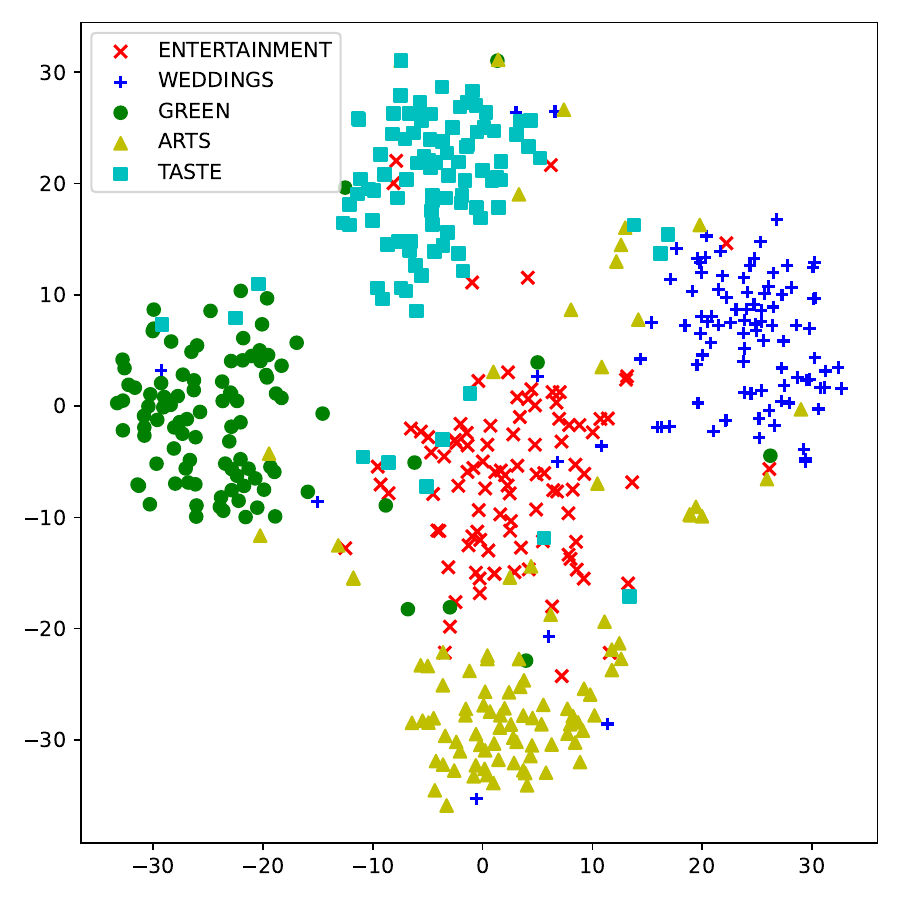}}
    \caption{The t-SNE figure for ContrastNet(a), LaSAML(b), Ours(c). 100 samples from 5 classes are sampled in one testing episode (N=5,K=1).}
    \label{figure:embedding_visualization}
\end{figure}

\noindent
\textbf{Visualization}
We use t-SNE \cite{maatenVisualizingDataUsing2008} to visualize our learned embeddings, as done by \cite{luoDonMissLabels2021} and \cite{chenContrastNetContrastiveLearning2022}. The model trained on 1-shot tasks is chosen instead of the 5-shot tasks to better demonstrate its learning and generalization abilities. ContrastNet and LaSAML, two strong methods, are chosen as our baselines for comparison. The results shown in Figure \ref{figure:embedding_visualization} demonstrate that our embeddings are more discriminative between different classes and more compact within the same class compared to ContrastNet. LaSAML shares a similar trend with our label template method, but our embeddings display more evenly distributed class representations and clearer boundaries between classes. This indicates our model better captures intra-class and inter-class semantic differences, as reflected in Table \ref{table:our_result}.

\section{Conclusion}
In this paper, we put forward a novel meta-learning framework for FSTC task. With the guidance of the label template, our method employs the supervised contrastive learning and attention mechanism to enhance the capability of the Prototypical Network. The framework can acquire more typical and robust feature vector for FSTC tasks with just few iterations. Extensive experiemntal results confirm the superiority of our method over traditional prototype networks and most of strong baselines, notably in the 1-shot scenario. In upcoming research, we aim to expand the proposed framework to include a wider range of few-shot tasks.\begin{credits}
    \subsubsection{Acknowledgments} This work was supported by the National Natural Science Foundation of China under Grants 62101079, 62072062 and U20A20176; the Natural Science Foundation of Chongqing under Grant cstc2021jcyj-msxm0755; the Venture and Innovation Support Program for Chongqing Overseas Returnees under Grant cx2021012; the Joint Fund of Ministry of Education of China for Equipment Pre-Research under Grant 8091B032127.
\end{credits}

\bibliographystyle{splncs04}
\bibliography{9589}\end{document}